\documentclass[11pt]{article}
\pdfoutput=1

\usepackage[margin=1in]{geometry}
\usepackage[T1]{fontenc}
\usepackage[utf8]{inputenc}
\usepackage{times}
\usepackage{microtype}
\usepackage{graphicx}
\usepackage{subcaption}
\usepackage{float}
\usepackage[section]{placeins}
\usepackage[absolute,overlay]{textpos}
\usepackage{booktabs}
\usepackage{amsmath}
\usepackage{hyperref}
\usepackage[english]{datetime2}
\usepackage[numbers]{natbib}
\usepackage{authblk}
\usepackage[most]{tcolorbox}
\DTMnewdatestyle{monthyeardate}{%
}

\title{Interpretable GOHR Agents via Sparse Autoencoders}
\author[1]{Shiwei Tan}
\author[1]{Yusong Zhao}
\author[1]{Weiyi Qin}
\author[1]{Wentian Wang}
\author[1]{Jacob Feldman} 
\author[1]{Lazaros K. Gallos} 
\author[1,3]{\\Paul B. Kantor}
\author[1]{Vladimir Menkov}
\author[1,2]{Hao Wang}
\affil[1]{Rutgers University}
\affil[2]{University of Illinois Urbana-Champaign}
\affil[3]{Paul B Kantor Consultant}
\date{\today}

\begin{document}
\begin{textblock*}{7cm}(12cm,1cm)
\raggedleft
\small
\textbf{TECHNICAL REPORT}\\
\DTMsetdatestyle{monthyeardate}\today
\end{textblock*}

\maketitle

\begin{abstract}
A central challenge in interpreting learned decision-making systems is to determine whether their internal representations contain concepts that help explain their behavior.
We report interpretability experiments for a tokenized autoregressive Transformer agent in the Game of Hidden Rules (GOHR).  We focus on a compact two-rule task in which both hidden rules map object shapes to target buckets, but with different permutations.  The policy is trained on episodes sampled from these two hidden rules and then evaluated with fixed weights.  It is never given a rule label and does not use an explicit rule classifier; any rule information must be inferred implicitly from interaction history.  In this setting, the correct rule is not identifiable before the agent tries an informative move and observes accept/reject feedback.  Sparse autoencoders (SAEs) trained on the agent's decision-token embeddings recover this structure.  When held-out decisions are labeled by simple concepts such as the chosen shape or bucket, SAE dimensions that are highly selective for a concept cover most decisions where that concept is present.  Individual SAE dimensions also correspond to interpretable strategies such as probing one rule hypothesis and switching after negative feedback.
\end{abstract}

\section{Introduction}
Let us suppose that an AI system, as it solves problems involving some kind of reasoning, formulates and then uses something that might be called ``concepts.'' Something like this is posited in the important paper by Olah et al. \cite{olahBuildingBlocksInterpretability2018}. Here, we report on a model study using a relatively small neural network and ask whether we can associate specific nodes or combinations of nodes with specific concepts. To do this, we selected a model environment in which it seems clear that (a) relevant concepts exist and (b) their presence can be demonstrated through the behavior of a small ``laboratory-scale'' learning network.

The Game of Hidden Rules (GOHR) is a controlled rule-learning environment in which an agent must infer a latent rule from sparse accept/reject feedback \citep{pulick_game_2022,pulick_comparing_2024}.  At each step, the agent observes a board containing multiple objects and chooses an object-bucket move.  The environment accepts or rejects the move according to a hidden rule, but the rule identity itself is never directly 
revealed. This makes GOHR useful for studying abstraction, trial-and-error learning, and interpretability in sequential decision-making systems.

Since the board and feedback do not reveal the rule  directly, the agent must infer which mapping is active from the interaction history. Although the two rules studied here depend only on object shape, each object is represented by multiple attributes, including shape, color, and position.  The agent therefore observes a richer board state than the rule actually requires.  Solving the task requires using feedback to infer the active shape-to-bucket mapping while ignoring attributes that are irrelevant in this setting.  This gives a controlled test of whether the learned representation separates behaviorally relevant concepts from other visible board features.

This report focuses on a compact two-rule setting where the learned behavior is easy to validate.  Both rules map object shapes to target buckets, but they use different shape-to-bucket permutations, chosen so that a shape-bucket move accepted by one rule is rejected by the other.  This gives a diagnostic case for policy interpretability: the relevant hidden variable is known to the experimenter, the expected adaptive strategy is simple, and the agent's behavior can be checked directly in rollouts.  A natural successful strategy is to try one plausible shape-to-bucket mapping, use accept/reject feedback as evidence, and switch mappings if the first hypothesis is contradicted.

There are two distinct questions in this case study.  The behavioral question is whether a trained, frozen policy uses interaction history to act as if it has inferred the current hidden rule.  The representational question is whether post hoc SAE analysis exposes internal features aligned with rule-relevant concepts and rule-conditioned action choices.  We evaluate both questions: first by checking for probe-and-switch behavior in rollouts, and then by asking whether SAE features recover behaviorally meaningful shape, bucket, and action-strategy structure.

\section{Method}

\begin{figure}[H]
\centering
\includegraphics[width=\linewidth]{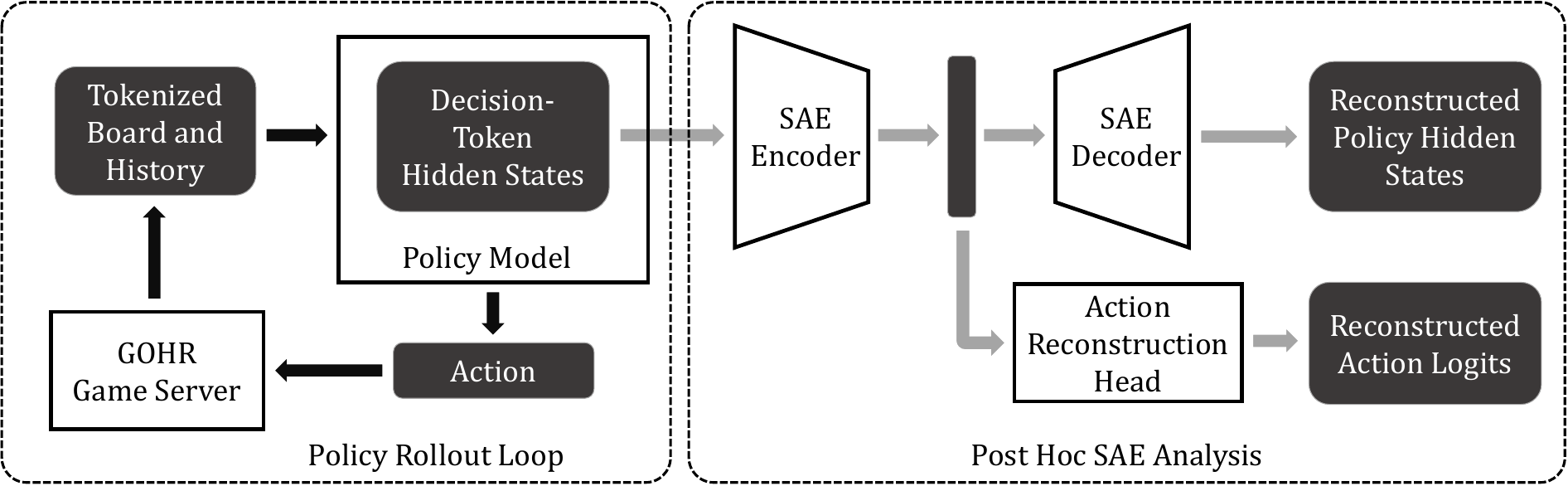}
\caption{Workflow overview.  The left panel shows the normal policy loop, in which the policy hidden states are used to choose actions and GOHR feedback updates the next state/history.  The right panel shows the post hoc SAE analysis: saved policy hidden states are encoded into a sparse representation, which is then decoded to reconstruct the hidden states and passed through an action-reconstruction head.}
\label{fig:workflow}
\end{figure}

Figure~\ref{fig:workflow} summarizes the separation between the policy rollout loop and the post hoc SAE analysis.  During rollout, the policy maps the current game state and interaction history to an action, and the GOHR server appends feedback to the next history.  The SAE analysis is performed afterward on saved policy hidden states; it does not participate in the agent's interaction with the environment.

\subsection{Task Setup}

Each GOHR board contains a set of colored geometric objects.  At every step, the agent chooses an object and assigns it to one of four buckets.  A hidden rule determines whether the move is accepted.  If the move is accepted, the selected object is cleared; if it is rejected, the object remains and the failed action becomes part of the interaction history.  The episode objective is to clear the board while making as few rejected moves as possible.

Formally, let \(\rho\in\{R_1,R_2\}\) denote the hidden rule for an episode.  The state \(s_t\) consists of the current board \(B_t\) and the previous action-feedback pairs \((a_i,f_i)_{i<t}\), where \(a_i=(o_i,k_i)\) selects object \(o_i\) and bucket \(k_i\), and \(f_i\in\{0,1\}\) is the accept/reject feedback.  The policy does not observe \(\rho\).  In the task studied here, both candidate rules are shape-to-bucket mappings, so the same object attribute is relevant under both rules, but the correct bucket for a given shape differs.

\begin{table}[H]
\centering
\small
\begin{tabular}{lcc}
\toprule
Object shape & Rule \(R_1\) bucket & Rule \(R_2\) bucket \\
\midrule
Circle & 0 & 1 \\
Star & 1 & 3 \\
Square & 2 & 0 \\
Triangle & 3 & 2 \\
\bottomrule
\end{tabular}
\caption{The two hidden mappings used in the main experiment, shown with anonymous rule names.  The agent observes only board objects and accept/reject feedback, not the active rule label.}
\label{tab:rule_mapping}
\end{table}

This setup creates a simple source of partial observability.  Before any feedback, both mappings in Table~\ref{tab:rule_mapping} are plausible.  A first move can therefore function as a probe: if the move is accepted, the agent can continue under that mapping; if it is rejected, the alternative mapping becomes the better hypothesis.  This probing structure is what we later look for in the policy representation.

\paragraph{Experiment protocol.}
The policy is trained before the interpretability analysis on episodes sampled from the two rules in Table~\ref{tab:rule_mapping}.  At the start of each episode, the environment samples one hidden rule; the agent observes the board and previous accept/reject feedback, but never observes the sampled rule label.  After training, the rollouts used in this report are generated with frozen policy weights.  There is no gradient update, no online learning, and no explicit rule-identification module during evaluation.  Thus, when the policy switches behavior after a rejection, that switch must be implemented by the Transformer through its tokenized interaction history.  The SAE is trained afterward on saved decision-token embeddings from these frozen rollouts, and is used only to analyze the policy representation.

Each episode starts with a nine-object board.  A perfectly successful episode therefore contains nine accepted moves, plus any rejected probes needed to disambiguate the rule; training and evaluation runs use a horizon of 100 attempted moves, but the rollouts analyzed here typically finish after the board has been cleared.

\subsection{Tokenized Autoregressive Agent}

The policy uses a tokenized state representation.  Board objects, previous actions, and feedback are serialized into one interaction history, allowing the Transformer to attend across the current board and the evidence accumulated so far.  For board state \(B_i\), the corresponding board block \(\mathcal{B}_i\) is serialized as object tokens bracketed by start and end markers:
\[
\mathcal{B}_i =
\langle\mathrm{start}\rangle\;[x]\;[y]\;[\mathrm{shape}]\;[\mathrm{color}]\;[\mathrm{size}]\;\cdots\;\langle\mathrm{end}\rangle .
\]
Each object contributes separate coordinate and attribute tokens.  Let \((u_i,v_i)\) denote the \(x\)- and \(y\)-coordinates selected on attempt \(i\), \(k_i\) the selected bucket, and \(f_i\) the resulting accept/reject feedback.  The completed attempt is then serialized as
\[
\mathcal{A}_i =
\langle x\_\mathrm{cls}\rangle\;[u_i]\;
\langle y\_\mathrm{cls}\rangle\;[v_i]\;
\langle \mathrm{bucket\_cls}\rangle\;[k_i]\;[f_i].
\]
For a decision at time \(t\), the full input sequence contains up to eight completed rounds followed by the current board and the next-coordinate query:
\[
\mathcal{B}_{t-m}\mathcal{A}_{t-m}\;\cdots\;\mathcal{B}_{t-1}\mathcal{A}_{t-1}\;\mathcal{B}_t\langle x\_\mathrm{cls}\rangle ,
\qquad m\le 8 .
\]
Both accepted and rejected moves are included in the history, so negative feedback can guide later decisions.

In the tokenized policy, the object \(o_t\) is selected by outputting its board coordinates.  For current state \(s_t\), the action \(a_t=((u_t,v_t),k_t)\) is generated autoregressively:
\[
\pi_\theta(a_t=((u,v),k)\mid s_t)
=P_\theta(u_t=u\mid s_t)
P_\theta(v_t=v\mid s_t,u_t=u)
P_\theta(k_t=k\mid s_t,u_t=u,v_t=v).
\]
This factorization makes the policy choose a board coordinate first and then choose a bucket for the selected object.  The hidden states at \(\langle x\_\mathrm{cls}\rangle\), \(\langle y\_\mathrm{cls}\rangle\), and \(\langle \mathrm{bucket\_cls}\rangle\) are saved as policy embeddings for downstream SAE analysis.  Table~\ref{tab:tworule} summarizes the resulting experimental setup.

\begin{table}[H]
\centering
\small
\begin{tabular}{ll}
\toprule
Field & Value \\
\midrule
Task & Two shape-to-bucket rules \\
Rule sampling & One hidden mapping per episode \\
Rule label & Never observed by the policy \\
Input sequence & Completed rounds followed by current-board query \\
Policy & 6-layer Transformer, \(d_{\mathrm{model}}=256\), 8 heads \\
History window & 8 rounds, max sequence length 512 \\
Training & SFT warm start followed by Advantage Actor-Critic (A2C) fine-tuning \citep{mnih2016asynchronous} \\
Evaluation & Frozen policy rollouts; no online updates \\
A2C horizon & 10,000 episodes, horizon 100 \\
\bottomrule
\end{tabular}
\caption{Core setup for the two-rule tokenized-agent experiment.  The two hidden mappings are shown explicitly in Table~\ref{tab:rule_mapping}.}
\label{tab:tworule}
\end{table}

\subsection{Sparse Autoencoder Analysis}

We train sparse bottleneck autoencoders on saved policy embeddings from the autoregressive decision tokens, following the general goal of dictionary-learning SAEs \citep{cunningham_sparse_2023}.  The encoder maps a policy embedding \(h\) to a nonnegative sparse code \(z\), and the decoder reconstructs the original embedding:
\[
z=E_\phi(h),\qquad \hat h=D_\psi(z),\qquad z\ge 0.
\]
In the run analyzed here, \(E_\phi\) maps each policy embedding through a ReLU sparse bottleneck, and \(D_\psi\) reconstructs the embedding from that bottleneck.  The training objective penalizes reconstruction error and latent activity and also includes an auxiliary loss from a linear action-reconstruction head operating on \(z\), encouraging the SAE dimensions to retain behaviorally relevant information.  Intuitively, an SAE dimension is useful if it activates on a small, coherent subset of decision states and the decoder can use it to reconstruct the policy representation.

The auxiliary head predicts the agent's policy logits from the same sparse code.  Let \(\ell_t\) denote the saved action-logit vector for the agent at decision state \(t\), and let
\[
\hat \ell_t=G_\eta(z_t)
\]
be the output of the action-reconstruction head.  The representation-learning terms in the SAE objective are
\[
\mathcal{L}_{\mathrm{SAE}}
=\|D_\psi(E_\phi(h_t))-h_t\|_2^2
+\beta\|z_t\|_1
+\alpha\|G_\eta(z_t)-\ell_t\|_2^2,
\]
with \(\alpha=0.1\) in the analyzed run.

We evaluate learned SAE features against human-defined concepts such as selected shape, selected bucket, and rule-conditioned action patterns.  For each binary concept \(c\), we estimate the precision of SAE dimension \(j\) as
\[
\mathrm{Prec}(j,c)=P(c=1\mid z_j>0).
\]
At threshold \(\tau\), the selected SAE feature group is \(D_c(\tau)=\{j:\mathrm{Prec}(j,c)\ge \tau\}\).  The reported recall asks whether at least one selected feature fires on decisions whose external concept label is positive:
\[
\mathrm{Recall}(c,\tau)=P\left(\exists j\in D_c(\tau): z_j>0 \mid c=1\right).
\]
Sweeping \(\tau\) produces a recall-vs.-precision-threshold curve.  A high curve indicates that concept-selective SAE features cover many concept-positive decisions even when we require each selected feature to be precise.

\section{Experiments}

\subsection{Supervised Warm Start}

The supervised warm start is an interface-learning stage before RL.  Its role is to make the autoregressive policy produce valid board coordinates and plausible bucket tokens, so that later A2C training can focus on feedback-driven rule inference rather than basic output formatting.  Unlike A2C, this stage does not let the current policy generate its own rollouts.  Instead, we first generate an offline set of two-rule episodes with a scripted data generator that proposes candidate moves and checks them against the GOHR environment.  Successful moves provide the supervised action targets.  Failed moves may still appear earlier in the same sequence, so the model sees them as feedback history, but the imitation loss is applied to the next successful action rather than to the failed action itself.  The model is then trained with teacher forcing on these program-generated action sequences, with the rule identity still hidden from the policy.  This stage does not train a separate rule classifier and is not evaluated with the SAE.

We evaluate the warm start mainly by legal-coordinate accuracy.  Exact action matching is stricter: it requires the model to match the particular logged target action.  Because a board may contain several objects that are legal or strategically equivalent to move next, a prediction can be a valid coordinate even when it differs from the logged target and is therefore counted as an exact-action mismatch.  In the two-rule SFT run, legal-coordinate accuracy rapidly approaches 1.0, reaching approximately 0.999 by the end of training; Table~\ref{tab:sft_metrics} reports the final logged metrics.

\begin{table}[H]
\centering
\small
\begin{tabular}{lrrr}
\toprule
Stage & Exact acc. & Bucket acc. & Legal coord. acc. \\
\midrule
SFT final epoch & 0.436 & 0.576 & 0.9997 \\
\bottomrule
\end{tabular}
\caption{Supervised warm-start metrics from the final logged epoch.  Exact accuracy requires the predicted coordinate and bucket to match the program-generated target action.  Bucket accuracy evaluates only the predicted bucket.  Legal-coordinate accuracy measures whether the predicted coordinate selects a valid board object, which is the main interface-quality metric because multiple legal moves may be valid on the same board.}
\label{tab:sft_metrics}
\end{table}

\subsection{A2C Behavior}

After SFT, the policy is fine-tuned with Advantage Actor-Critic in the live two-rule environment.  During this training stage, the policy improves from reward feedback over many episodes drawn from both hidden rules.  During the evaluation rollouts used for interpretability, however, the weights are fixed.  Any within-episode adaptation must therefore arise from inference over the sequence history rather than from additional fine-tuning.

In the two-rule setting, the ideal policy may still make one initial exploratory mistake when the current board does not distinguish the two shape-to-bucket mappings.  With a nine-object board, this corresponds to an idealized error rate of about \(0.5/9 \approx 0.056\).  We therefore interpret rollouts through the expected default-and-switch strategy: early failed probes are compatible with the task ambiguity, while later moves should follow the inferred mapping.

The rollout pattern is easy to interpret.  The agent often tries one shape-to-bucket mapping first.  When that hypothesis is correct, the default strategy succeeds immediately.  When it is wrong, the first probe must be rejected because the two mappings assign different buckets to every shape; after observing the rejection, the agent can switch to the other mapping and complete the episode.  This is the expected pattern under a simple two-hypothesis strategy, and Figure~\ref{fig:episodes} shows concrete rollouts where the trained policy follows it.  The important point for the SAE analysis is that this behavior creates separable decision states: pre-feedback probes, post-rejection switches, and ordinary moves after the rule has been inferred.
\subsection{SAE Rule and Feature Recovery}

\begin{figure}[t]
\centering
\includegraphics[width=0.72\linewidth]{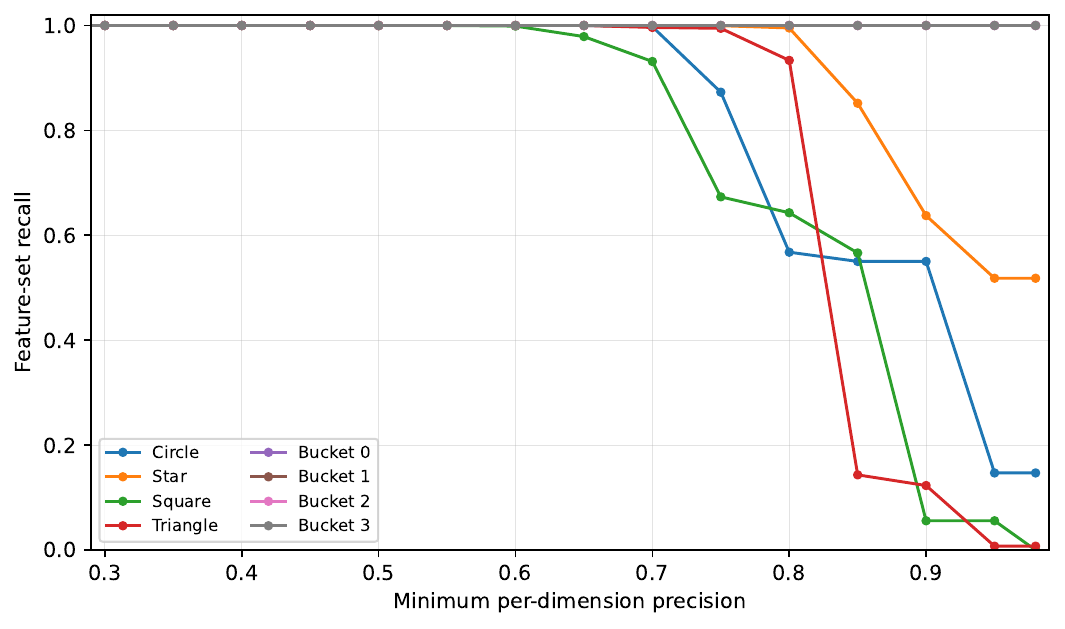}
\caption{Recall vs. required per-feature precision threshold for selected SAE feature groups.  For each human-defined concept, a point measures whether the union of all selected SAE dimensions activates when that concept is present.  The curves are computed over 10,000 held-out activation samples from the two-rule agent.  Differences among otherwise symmetric shape concepts can arise because the trained policy need not use a symmetric strategy: it may, for example, prefer clearing one shape before another, changing which concept-positive decision states appear and how often selected SAE features cover them.}
\label{fig:sae_curves}
\end{figure}

The SAE trained on decision-token embeddings provides the main quantitative interpretability result.  Figure~\ref{fig:sae_curves} shows that feature-set recall remains high even as the required per-dimension precision threshold increases.  Table~\ref{tab:sae_metrics} summarizes the same sweep numerically using the normalized area under the recall-vs.-precision-threshold curve:
\[
\mathrm{nAUC}(c)=\frac{1}{0.98-0.30}\int_{0.30}^{0.98}\mathrm{Recall}(c,\tau)\,d\tau .
\]
The nAUC value is close to one when the concept can be recovered across most precision requirements, and close to zero when high-precision dimensions cover only a small fraction of concept-positive examples.  The curve only reports recall.  Table~\ref{tab:sae_metrics} also reports the precision of the OR-combined feature group at \(\tau=0.8\), which can be lower than the precision of any individual selected dimension when many dimensions share positives but contribute different false positives.

\begin{table}[t]
\centering
\small
\begin{tabular}{lrrrr}
\toprule
Feature group & n & nAUC & Recall @ 0.8 & Group precision @ 0.8 \\
\midrule
Shape concepts & 4 & 0.824 & 0.785 & 0.722 \\
Bucket concepts & 4 & 1.000 & 1.000 & 0.513 \\
All concepts & 8 & 0.912 & 0.892 & 0.617 \\
\bottomrule
\end{tabular}
\caption{Aggregate SAE feature-matching metrics.  nAUC is the normalized area under the recall curve over precision thresholds 0.30--0.98.  Recall and group precision are reported at the representative precision threshold \(\tau=0.8\).  Group precision is \(P(\mathrm{concept}\mid\mathrm{any\ selected\ dimension\ active})\), where selected dimensions are combined with a logical OR.}
\label{tab:sae_metrics}
\end{table}
\FloatBarrier
  
\subsection{Episode-Level Visualizations}

\begin{figure}[ht]
\centering
\begin{subfigure}{0.95\linewidth}
  \centering
  \includegraphics[width=\linewidth]{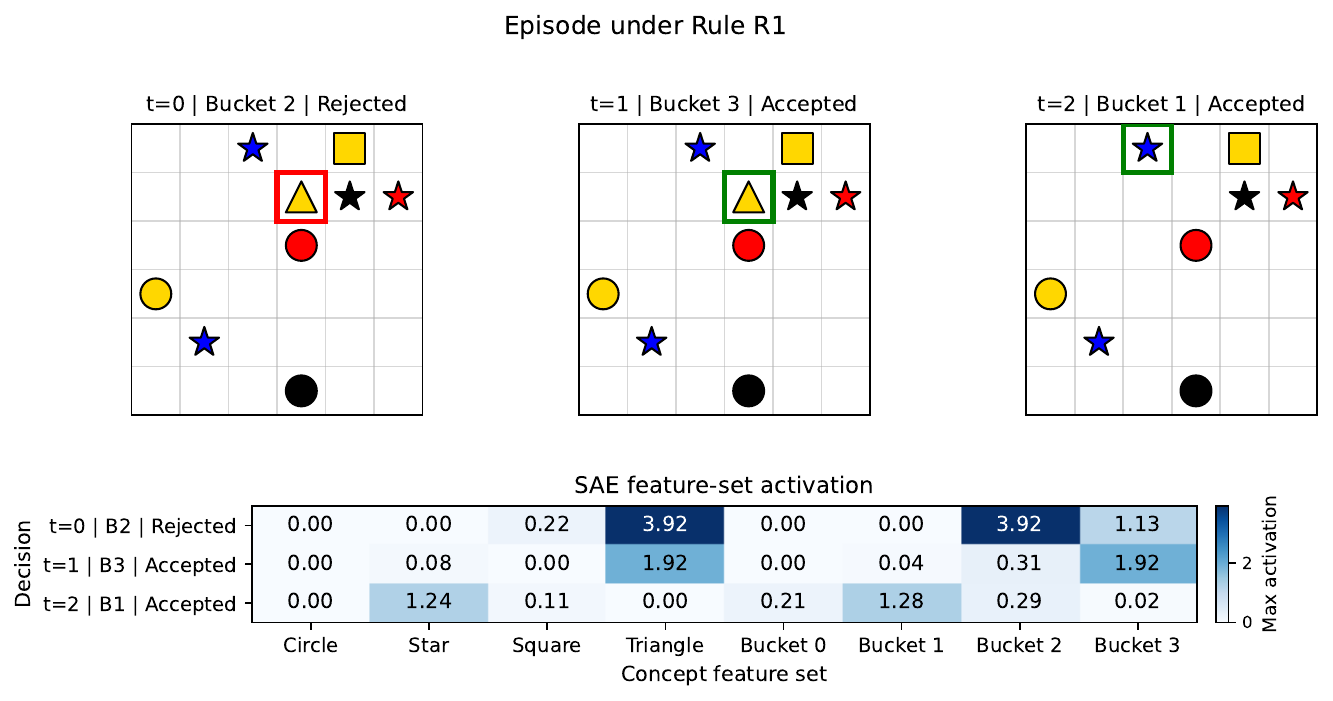}
  \caption{Probe, then switch after negative feedback.}
\end{subfigure}
\vspace{0.15em}
\begin{subfigure}{0.95\linewidth}
  \centering
  \includegraphics[width=\linewidth]{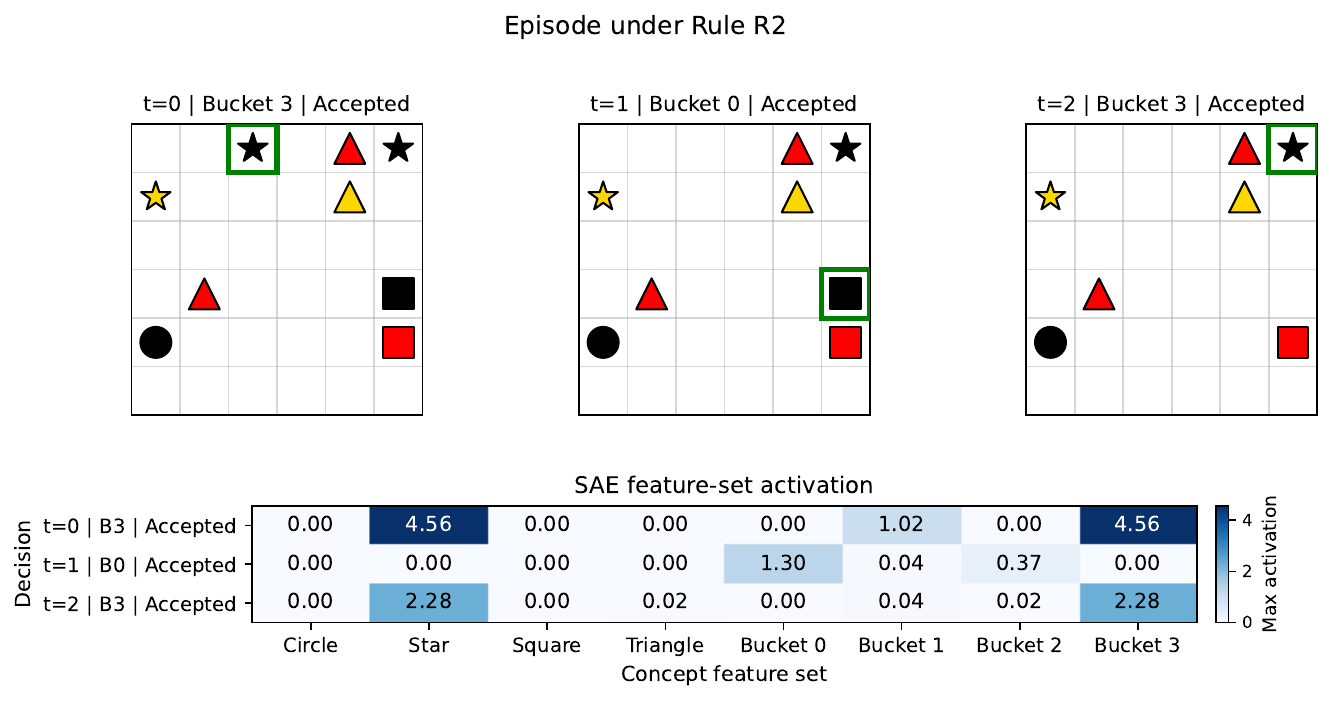}
  \caption{The same default strategy succeeds immediately.}
\end{subfigure}
\caption{Rule-conditioned behavior and corresponding SAE feature-set activations for the first three decisions of two episodes.  Red and green outlines mark rejected and accepted moves, respectively.  In each panel, the lower colored matrix is a heat map: each cell is the maximum activation among dimensions selected for the column concept at precision threshold \(0.75\), evaluated at the decision state named by the row label.}
\label{fig:episodes}
\end{figure}

The episode visualizations show the same structure in concrete rollouts.  In one rule condition, the agent begins with a move consistent with the alternate mapping; after it fails, later moves align with the active rule.  In the other condition, the same default strategy is already correct.  To connect these actions to the SAE representation, Figure~\ref{fig:episodes} includes a lower matrix in each panel showing a heat map over concept-specific SAE feature groups.  Rows correspond to the displayed decisions and columns correspond to shape or bucket concepts.  The cell corresponding to concept \(c\) and displayed decision \(t\) reports \(\max_{j\in D_c(0.75)} z_{t,j}\).  Under \(R_1\), the rejected triangle-to-bucket-2 probe jointly activates the Triangle and Bucket 2 feature sets; after feedback, the accepted triangle-to-bucket-3 move instead activates Triangle and Bucket 3.  Under \(R_2\), the corresponding shape and bucket feature sets are already jointly active on the first accepted move.  The selected SAE features therefore track the action components that change with the agent's rule-conditioned strategy.

\subsection{Interpretable Dimensions}

\begin{figure}[ht]
\centering
\begin{subfigure}{0.95\linewidth}
  \centering
  \includegraphics[width=\linewidth]{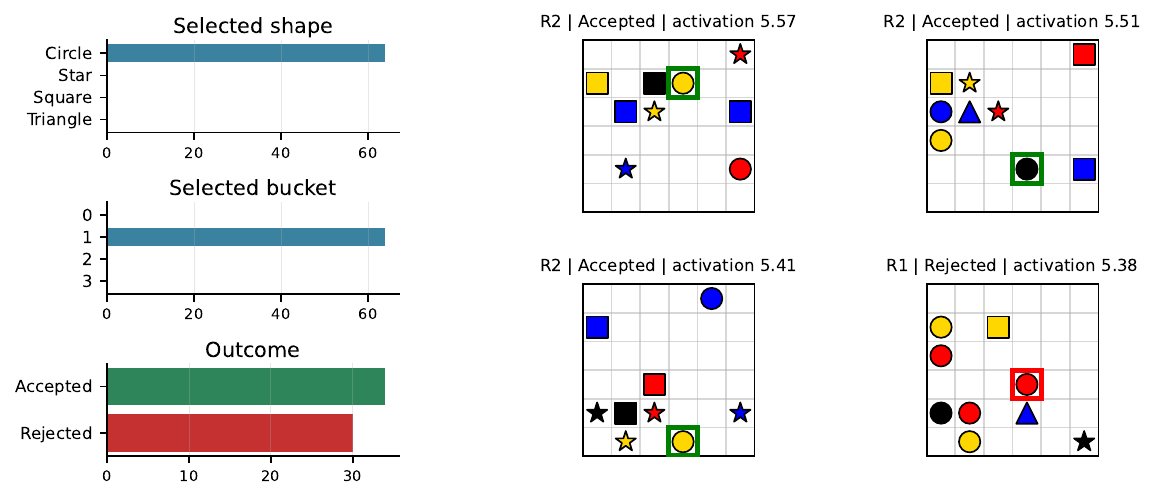}
  \caption{Dim 9697: circle-to-bucket-1 probes.}
\end{subfigure}
\vspace{0.35em}
\begin{subfigure}{0.95\linewidth}
  \centering
  \includegraphics[width=\linewidth]{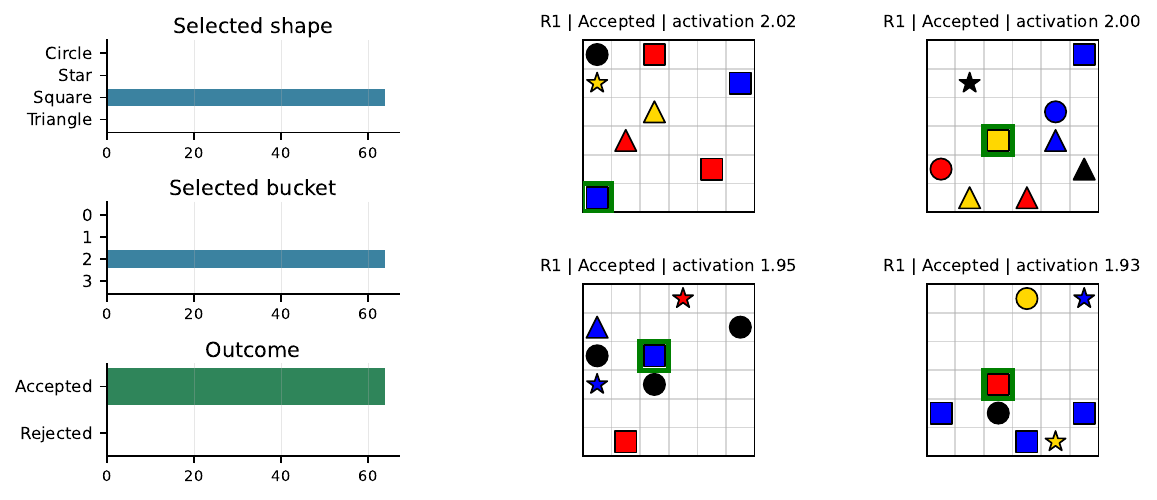}
  \caption{Dim 8720: successful square-to-bucket-2 moves.}
\end{subfigure}
\caption{Examples of individual SAE dimensions with clean behavioral interpretations.  Each behavior profile aggregates the 64 highest-activation decisions for that dimension; the four boards show the highest-activation examples.}
\label{fig:dims}
\end{figure}

Several individual SAE dimensions have simple behavioral descriptions, as shown in Figure~\ref{fig:dims}.  For example, one dimension activates on decisions at the beginning of episodes in which the agent selects circles and sends them to bucket 1; these moves succeed about half the time.  In Table~\ref{tab:rule_mapping}, this move is correct under Rule \(R_2\) and incorrect under Rule \(R_1\), so its activation pattern matches the strategy of trying one rule hypothesis before the rule has been disambiguated.  Another dimension activates on successful moves under the other rule, in which the agent selects squares and sends them to bucket 2; these moves are consistently correct.  These examples are useful because they connect a latent activation, a concrete action, and the hidden-rule semantics in the same episode.

\FloatBarrier

\subsection{Scope}

The case study has three main limitations.  The environment uses two closely related shape-to-bucket rules, the rollout evidence is qualitative, and the feature-group metric can hide redundancy because multiple dimensions are combined with a logical OR.  These limitations restrict the strongest conclusion to the two-rule setting: in this setting, SAE features recover both aggregate task concepts and concrete rule-conditioned actions.  The setting is informative because it has a known latent structure and the recovered features can be checked against both concept labels and rollout behavior.

\section{Conclusion}

This report presents a compact SAE interpretability case study for a tokenized GOHR agent.  In a two-rule setting where the agent must use feedback to infer a hidden shape-to-bucket mapping, the learned policy exhibits a simple probe-and-switch strategy.  Post hoc sparse autoencoders trained on decision-token hidden states recover both aggregate task concepts and fine-grained action strategies in the policy representation.  The strongest evidence comes from combining three views of the same system: rollout behavior, concept-level feature recall, and individual SAE dimensions whose activations align with concrete rule-conditioned actions.

These results show that, in a controlled sequential decision-making task with known latent structure, SAE features can expose behaviorally meaningful internal structure rather than only static properties of the observed board.  This supports using small rule-learning environments as testbeds for interpretability methods before applying the same analyses to larger agents or more complex rule mixtures.  A natural next step is to report more detailed per-dimension statistics for the most interpretable features and to test whether the same approach scales to broader GOHR settings.

This kind of ``concept exposure'' may contribute to a better understanding of how a network system is ``reasoning'' and help advance alignment goals. However, we anticipate that ``black-box'' methods for extracting concepts from observed behavior may eventually be necessary as humans work to understand the behavior of frontier models and closed proprietary systems.

\section*{Acknowledgments}\label{sec:acknowledgments}

This work was part of a larger effort to determine whether current AI systems could solve GOHR tasks quickly enough to assist a human player. A2C systems such as the one used here do learn quickly enough to partner with a human. However, recent experiments, to be reported elsewhere, suggest that current LLM systems can deal with some classes of hidden rules about as well as humans can. 

This research is supported in part by the Defense Advanced Research Projects Agency under other agreement HR001124203; the content of this report  does not necessarily reflect the position or the policy of the Government; and no official endorsement should be inferred. Distribution Statement: Approved for public release; distribution is unlimited.

\bibliographystyle{plainnat}
\bibliography{custom}

\end{document}